\documentclass[a4paper,twoside]{article}
\usepackage[top=3.3cm,bottom=4.2cm,left=2.6cm,right=2.6cm]{geometry}
\usepackage{epsfig}
\usepackage{subcaption}
\usepackage{calc}
\usepackage{amssymb}
\usepackage{amstext}
\usepackage{amsmath}
\usepackage{amsthm}
\usepackage{multicol}
\usepackage{pslatex}
\usepackage{apalike}

\usepackage[dvipsnames]{xcolor}

\usepackage[labelfont=bf]{caption}
\usepackage[font=small,labelfont=bf]{caption}

\usepackage{color, colortbl}
\definecolor{Gray}{gray}{0.9}

\usepackage{pifont}
\usepackage{soul}
\newcommand{\cmark}{\ding{51}}
\newcommand{\xmark}{\textcolor{gray}{\ding{55}}}

\usepackage[pagebackref,breaklinks,colorlinks]{hyperref}

\usepackage[capitalize]{cleveref}
\crefname{section}{Sec.}{Secs.}
\Crefname{table}{Table}{Tables}
\crefname{table}{Tab.}{Tabs.}

\usepackage{array}
\newcolumntype{P}[1]{>{\centering\arraybackslash}p{#1}}
\newcolumntype{M}[1]{>{\centering\arraybackslash}m{#1}}

\usepackage{amsfonts}
\usepackage{mathtools}
\usepackage[ruled, lined, linesnumbered, commentsnumbered, longend, noend]{algorithm2e}
\UseRawInputEncoding
\usepackage{xcolor}

\usepackage{fancyhdr}
\usepackage{lastpage}
\usepackage{amsmath}
\usepackage{pdflscape}
\usepackage{breqn}

\usepackage{cuted}
\usepackage{capt-of}
\usepackage{amssymb}

\usepackage{algpseudocode}
\usepackage{makecell}
\setcellgapes{4pt}

\usepackage{algorithm2e}
\usepackage[bottom]{footmisc}
\usepackage{SCITEPRESS}     

\begin{document}
\title{Distortion-Aware Adversarial Attacks on Bounding Boxes \\of Object Detectors}
\author{\authorname{Pham Phuc}}
\author{\authorname{Pham Phuc\sup{1}, Son Vuong\sup{2,3}, Khang Nguyen\sup{4}, and Tuan Dang\sup{4}}
\affiliation{\sup{1} Ho Chi Minh City University of Technology, Vietnam, phuc.phamhuythien@hcmut.edu.vn}
\affiliation{\sup{2} VinBigData, Vietnam, v.sonvt15@vinbigdata.org}
\affiliation{\sup{3} VNU University of Engineering and Technology, Vietnam}
\affiliation{\sup{4} University of Texas at Arlington, USA, khang.nguyen8@mavs.uta.edu, tuan.dang@uta.edu}
}

\abstract{Deep learning-based object detection has become ubiquitous in the last decade due to its high accuracy in many real-world applications. With this growing trend, these models are interested in being attacked by adversaries, with most of the results being on classifiers, which do not match the context of practical object detection. In this work, we propose a novel method to fool object detectors, expose the vulnerability of state-of-the-art detectors, and promote later works to build more robust detectors to adversarial examples. Our method aims to generate adversarial images by perturbing object confidence scores during training, which is crucial in predicting confidence for each class in the testing phase. Herein, we provide a more intuitive technique to embed additive noises based on detected objects' masks and the training loss with distortion control over the original image by leveraging the gradient of iterative images. To verify the proposed method, we perform adversarial attacks against different object detectors, including the most recent state-of-the-art models like YOLOv8, Faster R-CNN, RetinaNet, and Swin Transformer. We also evaluate our technique on MS COCO 2017 and PASCAL VOC 2012 datasets and analyze the trade-off between success attack rate and image distortion. Our experiments show that the achievable success attack rate is up to $100$\% and up to $98$\% when performing white-box and black-box attacks, respectively. The source code and relevant documentation for this work are available at the following link: \href{https://github.com/anonymous20210106/attack_detector}{https://github.com/anonymous20210106/attack\_detector}.}
\maketitle \normalsize \setcounter{footnote}{0}

\section{Introduction}
Neural network-based detectors play significant roles in many crucial downstream tasks, such as 3D depth estimations \cite{dang2023multiplanar}, 3D point cloud registration \cite{nguyen2024real}, semantic scene understanding \cite{nguyen2024volumetric}, and visual SLAM \cite{dang2024v3d}. However, neural networks are proven to be vulnerable to adversarial attacks, especially for vision-based tasks. Starting from image classification, prior works \cite{DBLP:journals/corr/GoodfellowSS14,DBLP:conf/iclr/MadryMSTV18,moosavi-dezfooli_deepfool_2016} try to attack classification models systematically. Fast Gradient Sign Method (FGSM) \cite{DBLP:journals/corr/GoodfellowSS14} and Projected Gradient Descent (PGD) \cite{DBLP:conf/iclr/MadryMSTV18} leverage gradients of the loss function to add a minimal perturbation and find the direction to move from the current class to the targeted class. In this realm, DeepFool \cite{moosavi-dezfooli_deepfool_2016} formed this as an optimization problem to find both minimal distances and optimal direction by approximating a non-linear classification using the first order of Taylor expansion and Lagrange multiplier. Besides gradient-based approaches, \cite{alaifariadef} generated adversarial images by optimizing deformable perturbation using vector fields of the original one. Although adversarial attacks gain more attention and effort from researchers, crafting theories and practical implementation for this problem on object detectors is not well-explored compared to classification tasks.

\begin{figure*}
    \includegraphics[width=1.00\linewidth]{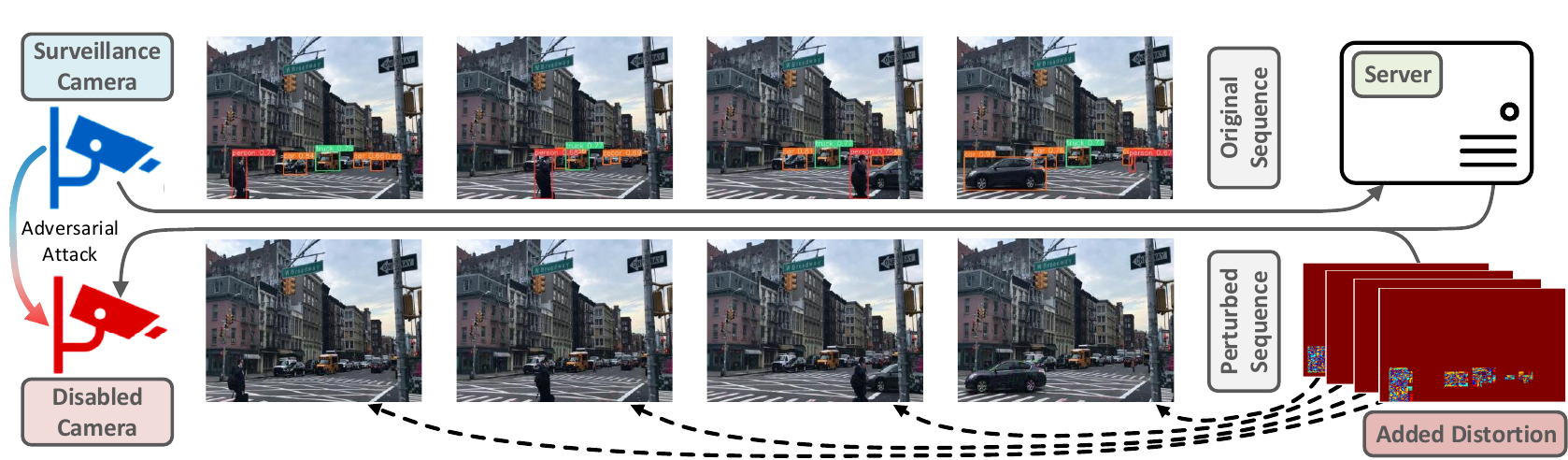}
    \captionof{figure}{Adversarial attacks on bounding boxes of an object detector with distortion awareness can perturb a sequence of images taken from a surveillance camera with a controllable added amount of distortion to obtain a certain success attack rate, making the object detector disabled. The demonstration video of the illustrated sequences is available at \href{https://youtu.be/y_sQqECMJIk}{YouTube}.}
    \vspace{-10pt}
    \label{fig:method_overview}
\end{figure*}

Motivated by adversarial attacks for classification, recent works \cite{song2018physical,lu_adversarial_2017,im2022adversarial,xie2017adversarial} attempt to perturb image detectors. Patch-based approach \cite{lu_adversarial_2017,song2018physical,DBLP:conf/aaai/LiuYLSCL19,du_physical_2022} adds random patches or human design patches into original images; these methods are reported to be effective in fooling the detectors, but the patches are apparently visible to the human eyes. Dense Adversary Generation (DAG) \cite{xie2017adversarial} considers fooling detectors as a fooling classifier for proposed bounding boxes: perturbing labels in each proposed bounding box to make the detector predict a different label other than the true one. Meanwhile, another method \cite{im2022adversarial} focuses on attacking location, objectness, or class confidence of YOLOv4 \cite{bochkovskiyyolov4} by noising the vehicle-related image using FGSM and PGD methods. However, this technique lacks the knowledge of individual bounding boxes, resulting in inaccurate added noises in multi-object images. Furthermore, the quality of adversarial images is not well-studied and is often disregarded in previous works due to their prioritization of attacking methods' effectiveness. 

Indeed, perturbing object detectors is far more challenging since the network abstracts location regression and object confidence, and loss functions are often multi-task learning. Self-exploring images and finding the best perturbation like \cite{moosavi-dezfooli_deepfool_2016,alaifariadef} for detectors become exhausting because of multi-task learning. As learning to detect objects in images heavily depends on the objective or loss functions, training detectors aims to minimize and converge these losses. Thus, one way to attack detectors is to increase losses for training samples to a certain level so that detectors misdetect or no longer recognize any objects. Through this observation, our approach is to find the optimal direction and distortion amount added to the targeted pixels with respect to these losses. Fig. \ref{fig:method_overview} demonstrates the practical application of our distortion-aware adversarial attack technique in real-world surveillance scenarios. The method introduces adversarial perturbations that cannot be recognized by humans but effectively disable object detection systems. It maintains a balance between preserving image quality and achieving high attack success rates, making it flexible across various practical situations. The unnoticeable nature of these distortions is crucial for adversarial use cases, as they remain visually undetectable while exploiting weaknesses in modern object detection models. This combination of stealth and effectiveness highlights the robustness of our approach.


To implement our method, we leverage the gradient from the loss function, like FGSM. While FGSM adds the exact amount of noise to every pixel except ones that do not change their direction, our approach uses magnitude from the gradient to generate optimal perturbations to all targeted pixels. As detectors propose bounding boxes and predict if objects are present in such regions before predicting which classes they belong to, object confidence plays an essential role in the detection task. We, therefore, inclusively use these losses and further sampling with a recursive gradient to take advantage of valuable information from all losses. We also find optimal perturbation amount iteratively as iterative methods produce better results than the fast methods. 

\renewcommand{\arraystretch}{1.5}
\begin{table}[h]
    \centering
    \resizebox{7.5cm}{!}{
    \begin{tabular}{m{6.3cm}| c c c}
        \hline
        {} & DAG & UEA & Ours \\
        \hline
        \large iterative added noises & \cmark & \xmark & \cmark \\
        \hline
        \large mostly imperceptible to human eyes & \cmark & \cmark & \cmark\\
        \hline
        \large distortion awareness & \xmark & \xmark & \cmark \\
        \hline
        \large stable transferability to backbones & \xmark & \xmark & \cmark \\
         \hline
        \large consistent with detection algorithms & \xmark & \xmark & \cmark \\
        \hline 
    \end{tabular}}
    \caption{Comparisons between our method and previous methods, including DAG \cite{xie2017adversarial} and UEA \cite{DBLP:conf/ijcai/WeiLCC19}, in terms of key properties.}
    \label{tab:key_properties}
    \vspace{-3pt}
\end{table}
\renewcommand{\arraystretch}{1}

In this work, our contributions are summarized as follows: (1) formalize a distortion-aware adversarial attack technique on object detectors, (2) propose a novel approach to attack state-of-the-art detectors with different network architectures and detection algorithms \cite{ren2015faster,lin2017focal,liu2021swin,Jocher_YOLO_by_Ultralytics_2023}, and (3) analyze and experiment our proposed technique on MS COCO 2017 \cite{lin2014microsoft} and PASCAL VOC 2012 \cite{everingham2015pascal} datasets with cross-model transferability and cross-domain datasets validation. Our key properties compared to previous methods \cite{xie2017adversarial,DBLP:conf/ijcai/WeiLCC19} are also shown in Tab. \ref{tab:key_properties}.


\vspace{-13pt}
\section{Related Work}

\indent \indent \textbf{Adversarial Attacks on Object Detectors}: Previous works in adversarial attacks on object detection can be categorized into optimization problems and Generative Adversarial Networks (GAN). The optimization problem is finding the adversarial images that satisfy the objective functions, while GAN generates adversarial images by training a generator that focuses on a classification or regression of the target network  \cite{DBLP:conf/ijcai/WeiLCC19}. Other methods use patches to fool the detectors \cite{song2018physical,du_physical_2022}, but noises are visible from a human perspective. We consider the adversarial attack as an optimization problem. Our method is conceptually similar to DAG \cite{xie2017adversarial}, but we focus more on finding the optimal direction and amplitude for each pixel to perturb given bounding boxes rather than drifting from one true class to another while proposing bounding boxes, which is impractical when class labels are unknown, especially in black-box attacks. Furthermore, we demonstrate the effectiveness of our methods on both one-stage and two-stage detectors. 

\textbf{Iterative Generative of Adversarial Images}: Inspired by the earliest study on classification problems \cite{DBLP:journals/corr/GoodfellowSS14}, the work \cite{kurakin2018adversarial} shows the effectiveness of iterative methods over one-shot methods by using the least-likely class method with FGSM to generate adversarial images for classification tasks. Another work \cite{alaifariadef} iteratively adds small deformation constructed by vector fields into images while DAG \cite{xie2017adversarial} performs iterative gradient back-propagation on adversarial labels for each target. Our method also uses iterative methods; however, it differs from those mentioned: we calculate the gradient over the iteratively permuted images and optimize this gradient under image distortion control. Moreover, we also focus on attacking general image detectors at different network architectures and detection methods, while \cite{DBLP:journals/corr/GoodfellowSS14,kurakin2018adversarial,alaifariadef} focus on attacking classifiers.

\textbf{Image Distortion Measurement}: Prior works \cite{kurakin2018adversarial,carlini2017towards,chen2018ead} used $l^{\infty}$ to measure the similarity between images original and adversarial images. $l^{\infty}$ effectively associates the corresponding features between pairs of images under changes, such as shifting or rotations \cite{wang2004image,lindeberg2012scale,rublee2011orb}. Regardless, as $l^{\infty}$ focuses on a per-pixel level, it lacks the illustration of how changes in a pixel might affect its neighboring pixels or might impact the overall pattern of the distorted image \cite{puccetti2023efficacy}. Other methods, such as mean square error (MSE), peak signal-to-noise ratio (PSNR), and contrast-to-noise ratio (CNR) are less sensitive to the human visual system \cite{lu_level_2019}. Therefore, we select Normalize Cross Correlation, which is robust to various image scales and less computational than Structural Similarity (SSIM) \cite{wang2004image} while maintaining the distortion imperceptibility to humans.

\vspace{-25pt}
\section{Problem Formulation}
\label{sec:formulation}
This section formalizes the attack strategy through key equations, including perturbation minimization (Eq. \ref{eq:image_minimizer}), discriminant functions (Eq. \ref{eq:attack_detection}), and optimization objectives (Eq. \ref{eq:detection_optimizer}).

\begin{figure}
    \centering
    \includegraphics[width=1.00\linewidth]{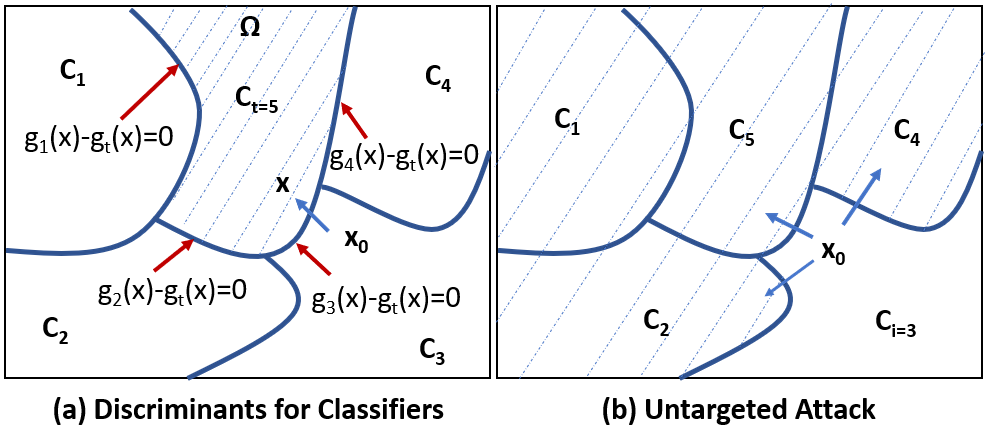}
    \caption{Illustration of adversarial attack with decision boundaries formed by k discriminant functions: attackers are looking for alternative $x$ that is similar to $x_0$ such that $g_i(x) < g_t(x_0)$ for $i=1,2,..,k$ and $t \neq i$ so that the model $f$ classify $x$ as $t$. An untargeted attack seeks $x$ such that the model, $f$, classifies $x$ as all $C_j$ where $i \neq j$. In this example, we choose $t=5$ and $i=3$.} 
    \label{fig:attacks}
\end{figure}

\subsection{Adversarial Attacks on Object Detectors}
\label{sec:formulation_definition}
\indent \indent \textbf{Defintion}: Let $\mathcal{I}$ be an RGB image of size of $m \times n \times 3$ with objects, $o_{1}, o_{2}, ..., o_{k}$, belonging to classes, $c_{1}, c_{2}, ..., c_{k}$. Similarly, the perturbed image is denoted as $\mathcal{I}^{'}$ but with the corresponding classes are now $c_{1}^{'}, c_{2}^{'}, ..., c_{k}^{'}$, where $\{c_{1}, c_{2}, ..., c_{k}\} \neq \{c_{1}^{'}, c_{2}^{'}, ..., c_{k}^{'}\}$. Therefore, our objective is to identify an algorithm such that the difference between $\mathcal{I}$ and $\mathcal{I}^{'}$ is minimized so that $\mathcal{I}^{'}$ can still perturb the detector, $f$, to misdetect objects but is mostly imperceptible to human eyes. The procedure, with $\varepsilon$ as the distortion (perturbation) amount, is written as:
\begin{equation}
    \underset{\varepsilon}{\text{minimize }}||\mathcal{I} - \mathcal{I}^{'}||, \text{ with } \varepsilon = \mathcal{I}^{'} - \mathcal{I}
    \label{eq:image_minimizer}
\end{equation}

\textbf{Discriminants for Classifiers}: Decision boundaries between a $k$-class-classifier are formed by $k$ discriminant functions, $g_{i}(\cdot)$, with $i = 1, 2, ..., k$, as illustrated in Fig. \ref{fig:attacks}a. Also, for untargeted attacks, misdetecting a particular object in $\mathcal{I}^{'}$ requires moving $f(o_{i})$ into a class other than its true class, $c_{i}^{'}$, as shown in Fig. \ref{fig:attacks}b. Thus, the domain, $\Omega_{i}$, that $f(o_{i})$ results in $c_{i}^{'} $ is defined as follows:
\begin{equation}
    \Omega_{i} \coloneqq \left\{o_{i} \mid g_{i}(o_{i}) - \underset{j \neq i}{\text{min}} \{g_{j}(o_{i})\} \leq 0 \right\}
    \label{eq:attack_classification}
\end{equation}

\textbf{Discriminants for Object Detectors}: Moreover, in the scope of object detection, accurate detections mainly rely on \textit{the class confidence scores of objects in bounding boxes} after non-max suppression. Therefore, the class confidence score should be inferred to be less than the confidence threshold for the detector to misdetect classes of objects in the bounding boxes. Reforming Eq. \ref{eq:attack_classification}, we obtain:
\begin{equation}
    \Omega_{i} \coloneqq \left\{ o_{i} \mid p(c_{i}) - \underset{j \neq i}{\text{min}} \{p(c_{j})\} \leq \mathcal{T} \right\}
    \label{eq:attack_detection}
\end{equation}
\begin{equation*}
    \text{ for } b_{i} \in \{b_{1}, ..., b_{k}\} \text{ and } \{b_{1}, ..., b_{k}\} \thicksim \{o_{1}, ..., o_{k}\} \text{ with }
\end{equation*}
$\thicksim$ represents the element-wise corresponding notation, $\{b_{1}, b_{2}, ..., b_{k}\}$ indicate the detected boxes in $\mathcal{I}$, $\mathcal{T}$ is the pre-defined confidence threshold, and $p(\cdot)$ represents the class probability function.

\subsection{Perturbing to Change Class Confidence Scores}
\indent \indent \textbf{Class Confidence Score}: To change the class confidence score of an object in a bounding box, we perturb its likelihood, $Pr(c_{i}\mid o_{i})$, to bring $p(c_{i})$ to be lower than the class probability, $p(c_{j})$, and the likelihood of another class, $Pr(c_{j}\mid o_{i})$, as follows:
\begin{equation}
    \begin{split}
        p(c_{i}) &= Pr(c_{i}\mid o_{i}) \cdot Pr(o_{i})\\
        & < Pr(c_{j}\mid o_{i}) \cdot Pr(o_{i}) = p(c_{j})
        \label{eq:confidence_score}
    \end{split}
\end{equation}

In short, to do this, the adversarial distortions should be added in each proposed bounding box. Therefore, based on Eq. \ref{eq:confidence_score} and $Pr(o_{i}) \geq 0$, meaning that there is a chance that the object is presented in the bounding box, Eq. \ref{eq:attack_detection} therefore can be rewritten into:
\begin{equation}
    \Omega_{i} \coloneqq \left\{ o_{i} \mid Pr(c_{i}\mid o_{i}) - \underset{j \neq i}{\text{min}} \{Pr(c_{j}\mid o_{i})\} \leq \mathcal{T} \right\}
    \label{eq:change_confidence_scores}
\end{equation}

\textbf{Objective Function for Object Detection}: Combining Eq. \ref{eq:image_minimizer} and Eq. \ref{eq:change_confidence_scores}, the generalized optimization generating an adversarial image that perturbs $f$ to misdetect $o_{i}$ in $b_{i}$ within $\mathcal{I}$ is defined as follows:
\begin{dmath}
    \underset{\varepsilon}{\text{minimize }} ||\mathcal{I}-\mathcal{I}^{'}|| \text{ \underline{such that} } \Omega_{i} \leq \mathcal{T}
    \label{eq:detection_optimizer}
\end{dmath}

\subsection{Perturbing through Detector Loss}
\indent \indent \textbf{Detector Loss}: Most commonly-used object detectors return predicted classes with their corresponding bounding box coordinates and confidence scores. In which, the loss function, $\mathcal{L}$, is the sum of classification loss, $\mathcal{L}_{cls}$, localization loss, $\mathcal{L}_{loc}$, and confidence loss, $\mathcal{L}_{obj}$, as below:
\begin{equation}
    \mathcal{L} =  \mathcal{L}_{loc} + \mathcal{L}_{obj} + \mathcal{L}_{cls}
    \label{eq:loss}
\end{equation}

\textbf{Perturbing through Detector Loss}: Based on Eq. \ref{eq:loss}, to \textit{desired target pixels} to perturb in an image, we add the amount of distortion as follows:
\begin{equation}
    \dfrac{\partial \mathcal{L}}{\partial \mathcal{I}} \cdot \mathbf{M}\left[f(\mathcal{I})\right]
    \label{eq:mask_out_cls_loss}
\end{equation}
where $\mathbf{M}$ represents all masks predicted by $f$ on $\mathcal{I}$, which is the sum of bounding boxes on an $m$-by-$n$ zeroes array, and $\partial$ indicates the partial derivative.

Therefore, to perturb classes' probabilities of an object in a bounding box, we can instead modulate it through the definition in Eq. \ref{eq:mask_out_cls_loss}, which effectively fools the object detectors during inference stages. The involvement of Eq. \ref{eq:mask_out_cls_loss} is shown in Eq. \ref{eq:update_adversarial_image} (Sec. \ref{sec:methodology_iterative_images}).

\vspace{-10pt}
\section{Methodology}
\label{sec:methodology}
We propose the white-box attack algorithm (Sec. \ref{sec:methodology_algorithms}) to find the most appropriate distortion amount, $\varepsilon$, via generating adversarial images, $\mathcal{I}^{'}$, iteratively (Sec. \ref{sec:methodology_iterative_images}) with distortion awareness (Sec. \ref{sec:methodology_distortion}). 

\subsection{Iterative Adversarial Images} 
\label{sec:methodology_iterative_images}
With the assumption that the object detector's network architecture is known, our method leverages the gradient of how pixels of predicted objects change when $\mathcal{I}$ passes through the network. In specific, we find the gradient ascent of targeted pixels to convert the original image, $\mathcal{I}$, to an adversarial image, $\mathcal{I}^{'}$. 

\textbf{Generating Iterative Adversarial Images}: However, the gradient derived from the total loss (Eq. \ref{eq:loss}) also gives the gradient of non-interested regions; meanwhile, we need to navigate the adversarial image to follow the gradient on specific bounding boxes. Using Eq. \ref{eq:mask_out_cls_loss}, we search for the adversarial image with respect to the gradient ascent of targeted pixels as follows:
\begin{equation}\
    \mathcal{I}_{i}^{'} = \mathcal{I}_{i-1}^{'} + \varepsilon = \mathcal{I}_{i-1}^{'} + \lambda \cdot \dfrac{\partial \mathcal{L}}{\partial \mathcal{I}_{i-1}^{'}}\cdot \mathbf{M}\left[f(\mathcal{I}_{i-1}^{'})\right]
    \vspace{-7.5pt}
     \label{eq:update_adversarial_image}
\end{equation}
\begin{equation*}
    \text{with } \mathbf{M}\left[f(\mathcal{I}_{i-1}^{'})\right] = \textbf{0}_{m \times n} + \sum_{i = 1}^{k} b_{i} \text{ and } \mathcal{I}_{0}^{'} = \mathcal{I}
\end{equation*}
where the subscripts, $i$ and $i-1$, denote current and previous iterations, respectively, $+$ sign denotes the gradient direction (ascending), $\lambda$ is the gradient ascent's step size, and $\mathbf{M}$ represents the aggregation of masks predicted by $f$ on $\mathcal{I}_{i-1}^{'}$, which is the sum of bounding boxes on an $m$-by-$n$ zero array.

\textbf{Distortion as Control Parameter}: Iterating Eq. \ref{eq:update_adversarial_image} over a considerable iterations, the generated adversarial image, $\mathcal{I}^{'}$, might get over-noised, which dissatisfies Eq. \ref{eq:image_minimizer} and eventually Eq. \ref{eq:detection_optimizer} regarding minimizing $\varepsilon$. Therefore, we introduce two strategies to control the distorted images:
\begin{equation}
    \mathcal{I}^{'} = 
    \begin{cases}
        \mathcal{I}_{i}^{'}, \quad \text{ if } \mathcal{D}(\mathcal{I}, \mathcal{I}_{i}^{'}) \geq \mathcal{S} \text{ or } f(\mathcal{I}_{i}^{'}) \geq \mathcal{R}\\
        \mathcal{I}_{i+1}^{'}, \text{ otherwise (using Eq. \ref{eq:update_adversarial_image})}
    \end{cases}
    \label{eq:update_with_control}
\end{equation} 
where $\mathcal{D}(\mathcal{I}, \mathcal{I}_{i}^{'})$ computes the distortion amount, $\varepsilon$, between $\mathcal{I}$ and $\mathcal{I}_{i}^{'}$ as subsequently defined in Eq. \ref{eq:distortion} (Sec. \ref{sec:methodology_distortion}), and $\mathcal{S}$ and $\mathcal{R}$ are the target distortion amount and the desired success attack rate, respectively, which are variants of $\mathcal{T}$.

\textbf{Differences of Proposed Strategies}: Both conditional statements of Eq. \ref{eq:update_with_control} eventually help Eq. \ref{eq:update_adversarial_image} to find the smallest iteration without brute-forcing over a larger iteration. Yet, the main difference between these equations is that $\mathcal{D}(\cdot, \cdot) \geq \mathcal{S}$ focuses on adding a desired distortion in the original image. Meanwhile, $f(\cdot) \geq \mathcal{R}$ concentrates on the desired success attack rate. Eq. \ref{eq:update_adversarial_image} is the extended version applied for detectors derived from \cite{kurakin2018adversarial}.

Furthermore, Eq. \ref{eq:update_adversarial_image} and Eq. \ref{eq:update_with_control} are explained on how they are involved in Alg. \ref{alg:adversarial_generation} (Sec. \ref{sec:methodology_algorithms}).

\setlength{\textfloatsep}{3.5pt}
\begin{algorithm}[t]
    \caption{Adversarial Images with Iterative Generation}
    \label{alg:adversarial_generation}
    \begin{small}
        \DontPrintSemicolon
        \SetKwInOut{KwIn}{Input}
        \SetKwInOut{KwOut}{Output}
        \SetKwFunction{FMain}{generator}
        \SetKwProg{Pn}{function}{}{}
        \KwIn{\mbox{$\mathcal{I} \coloneqq$ raw image, $\lambda \coloneqq$ step size}\\ 
        \mbox{$f \coloneqq$ detection model, $N\coloneqq$ max iteration}\\
        \mbox{$\mathcal{T} = \{\mathcal{S} \mid \mathcal{R}\} \coloneqq$ control parameters}}
        \KwOut{$\mathcal{I}^{'} \coloneqq$ adversarial image}
        \Pn{\FMain{$I, \lambda, f,  N, \{\mathcal{S} \mid \mathcal{R}\}$}}{
            $i = 0$, $\mathcal{I}_{i}^{'} = \mathcal{I}$\\
            $\{b_{1}, b_{2}, ..., b_{k}\} = \mathbf{B}\left[f(\mathcal{I})\right]$\\  
            \While{$i < N \textbf{ and } \{b_{1}, b_{2}, ..., b_{k}\} \neq \emptyset$}{
                $\mathbf{M} = \textbf{0}_{m \times n}$\\
                $\{b_{1}^{'}, b_{2}^{'}, ..., b_{k}^{'}\} = \mathbf{B}\left[f(\mathcal{I}_{i}^{'})\right]$\\
                \For{$ b_{j}^{'} \in \{b_{1}^{'}, b_{2}^{'}, ..., b_{k}^{'}\}$}{
                    \If{$\mathcal{D}(\mathcal{I}, \mathcal{I}_{i}^{'}) \geq \mathcal{S}$ \textbf{ or } $f(\mathcal{I}_{i}^{'}) \geq \mathcal{R}$}
                    {\textbf{break}}
                    $\mathbf{M} \leftarrow \mathbf{M} + b_{j}^{'}$
                }
                $I_{i+1} = I_{i} + \lambda \cdot \dfrac{\partial \mathcal{L}}{\partial \mathcal{I}_{i}^{'}}\cdot \mathbf{M}\left[f(\mathcal{I}_{i}^{'})\right]$\text{\quad(Eq. \ref{eq:update_adversarial_image})}\\
                $i \leftarrow i + 1$ \\
            }
            $\mathcal{I}^{'} = \mathcal{I}_{i}^{'}$\\
            \KwRet{$\mathcal{I}^{'}$}
        }       
    \end{small}
\end{algorithm}

\subsection{Normalized Cross Correlation}
\label{sec:methodology_distortion}
As Normalized Cross Correlation ($\texttt{NCC}$) depicts abrupt changes of targeted pixels to the average value of all image pixels while computing the similarity between two input images, we use $\texttt{NCC}$ for our work, as shown in Eq. \ref{eq:ncc_formula}. 
\begin{equation}
    \small
    \texttt{NCC}(\mathcal{I}, \mathcal{I}^{'}) = \dfrac{\sum_{i=1}^{n} \left(\mathcal{I}_{(i)} - \overline{\mathcal{I}}\right) \left(\mathcal{I}_{(i)}^{'} - \overline{\mathcal{I}^{'}}\right)}{\sqrt{\sum_{i=1}^{n} \left(\mathcal{I}_{(i)} - \overline{\mathcal{I}}\right)^2} \sqrt{\sum_{i=1}^{n} \left(\mathcal{I}_{(i)}^{'} - \overline{\mathcal{I}^{'}}\right)^2}}
    \label{eq:ncc_formula}
\end{equation}
with  $n$ is the number of pixels in $\mathcal{I}$ and $\mathcal{I}^{'}$, $\mathcal{I}_{(i)}$ indicates the $i^{\text{th}}$ pixel of $\mathcal{I}$, and $\overline{\mathcal{I}}$ represents the mean value of $\mathcal{I}$. 

Since $\texttt{NCC}(\mathcal{I}, \mathcal{I}^{'}) \in \left[0, 1\right]$ measures the similarity score between $\mathcal{I}$ and $\mathcal{I}^{'}$, we define the distortion metric (dissimilarity), $\mathcal{D}$, as the complement of $\texttt{NCC}$ in 1, as follows:
\begin{equation}
    \mathcal{D}(\mathcal{I}, \mathcal{I}^{'}) = 1 - \texttt{NCC}(\mathcal{I}, \mathcal{I}^{'})
    \label{eq:distortion}
\end{equation}

\subsection{Algorithm}
\label{sec:methodology_algorithms}
We propose the algorithm to find an adversarial image, $\mathcal{I}^{'}$, with an embedded perturbation amount, $\varepsilon$, from a raw RGB image, $\mathcal{I}$, under the control of the successful rate, $\mathcal{R}$, or the added distortion amount, $\mathcal{S}$. 

As illustrated in Alg. \ref{alg:adversarial_generation}, the algorithm first takes the bounding boxes that predicted objects provided by $f$ on a raw image $\mathcal{I}$. Hence, the adversarial image generation takes place iteratively until the predefined maximum iteration, $M$, is reached or no bounding boxes on $\mathcal{I}_{i}^{'}$ are detected by $f$. As the bounding boxes are re-predicted in each iteration, $\varepsilon$ is added based on the change of $\mathcal{L}$ with respect to the pixel's gradient ascent of $\mathcal{I}_{i}^{'}$. Using on Eq. \ref{eq:update_adversarial_image}, $\varepsilon$ is only added on the aggregated masks, $\mathbf{M}\left[f(\mathcal{I}_{i}^{'})\right]$, of the bounding boxes. To better control either $\varepsilon$ to be added or the success attack rate, $\mathcal{R}$, we also check if $\mathcal{D}(\mathcal{I}, \mathcal{I}_{i}^{'})$ or $f(\mathcal{I}_{i}^{'})$ exceeds the predefined threshold (Eq. \ref{eq:update_with_control}) to maintain the adversarial image to be adequately controlled; otherwise, $\varepsilon$ are kept adding in the next iteration. Note that the conditional statements in Alg. \ref{alg:adversarial_generation} can be used independently, which either controls $\mathcal{R}$ or $\mathcal{S}$. The analyses on control of $\mathcal{R}$ and $\mathcal{S}$ with respect to $\mathcal{I}_{i}^{'}$ are further provided in Sec. \ref{sec:analyses}.

\vspace{-15pt}
\section{Analyses}
\label{sec:analyses}
To verify our method's attacking feasibility, we analyze it with a subset of images on the \textit{most recent} detection models (YOLOv8 -- with various sizes).

\subsection{Convergence of Losses}
\label{sec:analyses_losses}
\begin{figure}[t]
    \centering
    \includegraphics[width=1.00\linewidth]{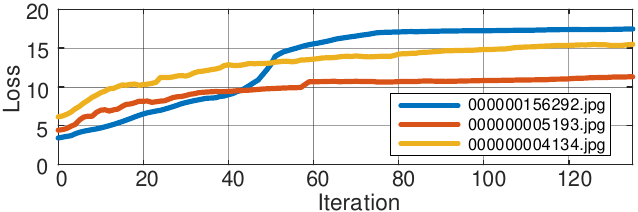}
    \caption{The convergence of loss over 120 iterations on a subset of images from the MS COCO 2017 dataset.} 
    \vspace{3pt}
    \label{fig:loss}
\end{figure}

The total loss consistently converges as adversarial images are iteratively generated, as shown in Fig. \ref{fig:loss}. To validate this behavior, we conducted extensive testing on numerous images from the MS COCO 2017 dataset, confirming that the convergence trend is consistent across the entire dataset. For visualization purposes, we randomly selected three representative images to illustrate this trend. Through our experiments, we found that 120 iterations strike an optimal balance between computational efficiency and attack performance, allowing sufficient time for the total loss to converge. This iteration count ensures that the results are representative and practical for attacks. 

This also shows that Alg. \ref{alg:adversarial_generation} can find adversarial images that can fool the object detectors. Also, the distortions of the adversarial images become larger as the iterations increase. Therefore, if we pick a recursively adversarial image before the convergence, we get a less-distorted image but eventually sacrifice the effectiveness of our attack. 

\vspace{-5pt}
\subsection{Image Distortion for Different Models with Confidence Thresholds and Success Rates}
\begin{figure}[h]
    \centering
    \includegraphics[width=1.00\linewidth]{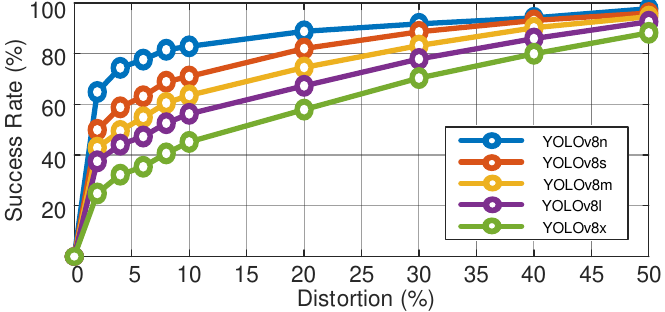}
    \caption{Relationship between attacking rate and target distortion on detection models set with confidence thresholds of $0.75$.} 
    \label{fig:NCC_objects}
\end{figure}

\indent \indent \textbf{Success Rates}: Fig. \ref{fig:NCC_objects} shows that YOLOv8n is the most vulnerable model with the least distorted image. In contrast, YOLOv8x is the hardest to attack, and its adversarial images are the most distorted compared to other models. Indeed, we can achieve a success attack rate of more than $80\%$ if image distortion is set by $10\%$. However, if the distortion rate increases from $10\%$, the attacking rate increases slowly. Overall, we can obtain a decent attacking rate by distorting only parts of images.

\begin{figure}[t]
    \centering
    \includegraphics[width=1.00\linewidth]{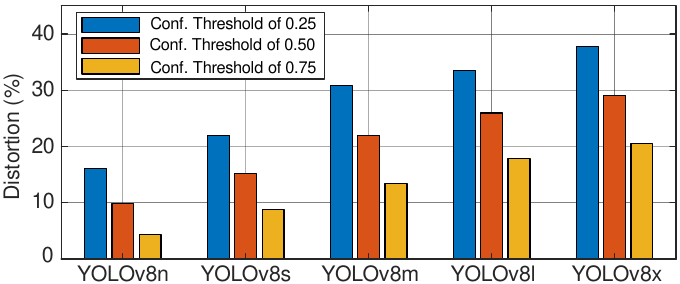}
    \caption{Relationship between confidence score and distortion at a success attack rate of $97\%$ on various-sized YOLOv8 models.} 
    \vspace{3pt}
    \label{fig:NCC_conf}
\end{figure}

\textbf{Confidence Scores}: We also evaluate Alg. \ref{alg:adversarial_generation} to see how average image distortion changes for each model when obtaining a desired attacking rate. Fig. \ref{fig:NCC_conf} depicts that attacking models with lower confidence scores causes the original images to be distorted more than the same model set with higher confidence scores. 

\subsection{Distortion Amount and Number of Iterations to Fool Different-Sized Models}
\label{sec:analyses_distortion_iterations}
\begin{figure}[t]
    \centering
    \includegraphics[width=1.00\linewidth]{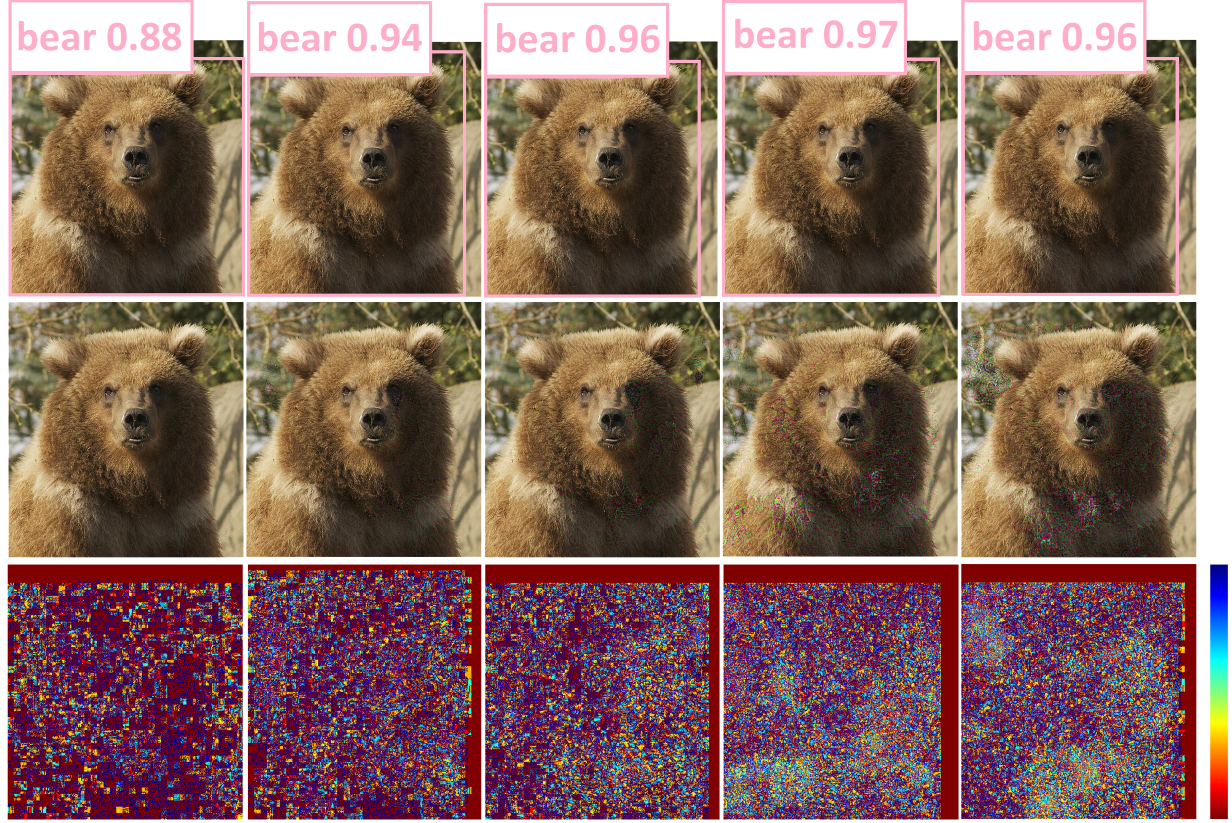}
    \caption{Comparisons between \textit{added distortion amounts (bottom row) on bounding box regions} to fool YOLOv8 from the smallest to the largest size, respectively (by column). Similarity scores computed by Eq. \ref{eq:ncc_formula} between original and perturbed images are $0.9996$, $0.9962$, $0.9925$, $0.9580$, and $0.9436$, respectively.} 
    \label{fig:yolo_model_diff}
    \vspace{6pt}
\end{figure}

\indent \indent \textbf{Distortion Amount}: The bottom row of Fig. \ref{fig:yolo_model_diff} shows the added distortion amounts (\textit{top row}) to generate the adversarial images (\textit{middle row}) among various-sized models. We notice that, for larger-sized models, our method tends to add more noise to prevent these models from extracting the objects' features and thereafter recognizing them, and vice versa. In this case, the features of the bear are perturbed. Another noticeable point is that the added distortion amount becomes more visible to human eyes when fooling the large-sized models, as depicted in the adversarial images and the heatmaps in the last two columns of Fig. \ref{fig:yolo_model_diff}. 

\begin{figure}[h]
    \centering
    \vspace{-3pt}
    \includegraphics[width=1.00\linewidth]{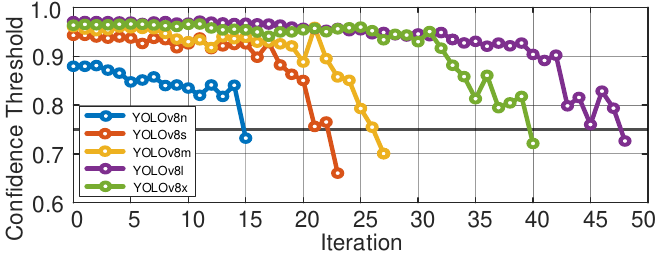}
    \caption{The minimum required iterations for Alg. \ref{alg:adversarial_generation} to fool various-sized YOLOv8 models with confidence thresholds of $0.75$.} 
    \label{fig:iterative_and_conf}
\end{figure}

\textbf{Number of Iterations}: As proven that our method needs more iterations to generate noises to fool large models, we provide the number of iterations to generate such perturbations, as shown in Fig. \ref{fig:iterative_and_conf}, which shows the approximately-proportional trend between the number of iterations to model sizes. 

\begin{figure*}[t]
    \centering
    \includegraphics[width=1.00\linewidth]{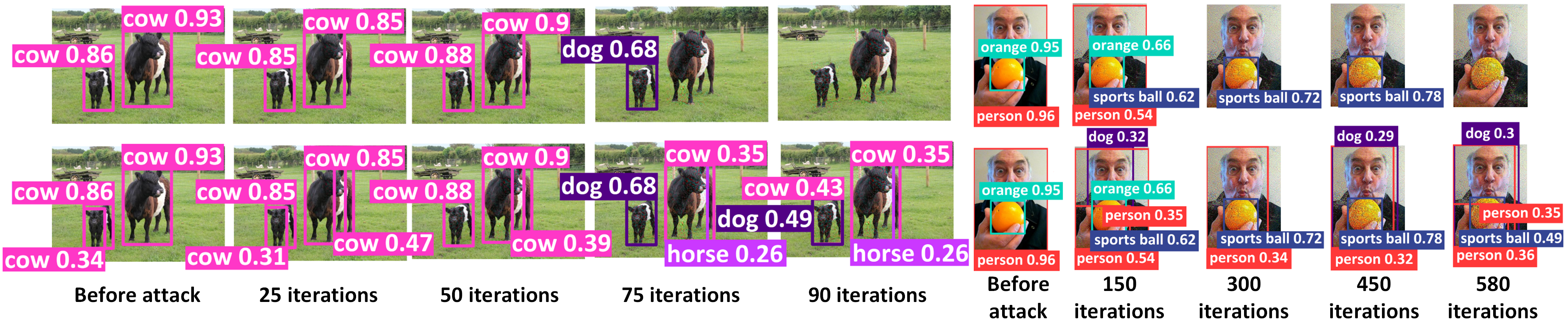}
    \caption{Adversarial images generated by Alg. \ref{alg:adversarial_generation} at different iterations and \textbf{how they affect the detector's performance} at confidence thresholds of \textbf{0.50 (top)} and \textbf{0.25 (bottom)}, respectively. The case of non-overlapping bounding boxes (\textit{left}) effectively causes the detector to recognize the wrong objects before misdetecting objects at the 90$^{th}$ iteration at a confidence threshold of $0.50$. Compared to the case where overlapped bounding boxes exist (\textit{right}), Alg. \ref{alg:adversarial_generation} takes more iterations (580 iterations) to fool the detector with the same configuration.}
    \vspace{-10pt}
    \label{fig:nonoverlap_overlap_attacks}
\end{figure*}

\begin{figure}[t]
    \centering
    \includegraphics[width=1.00\linewidth]{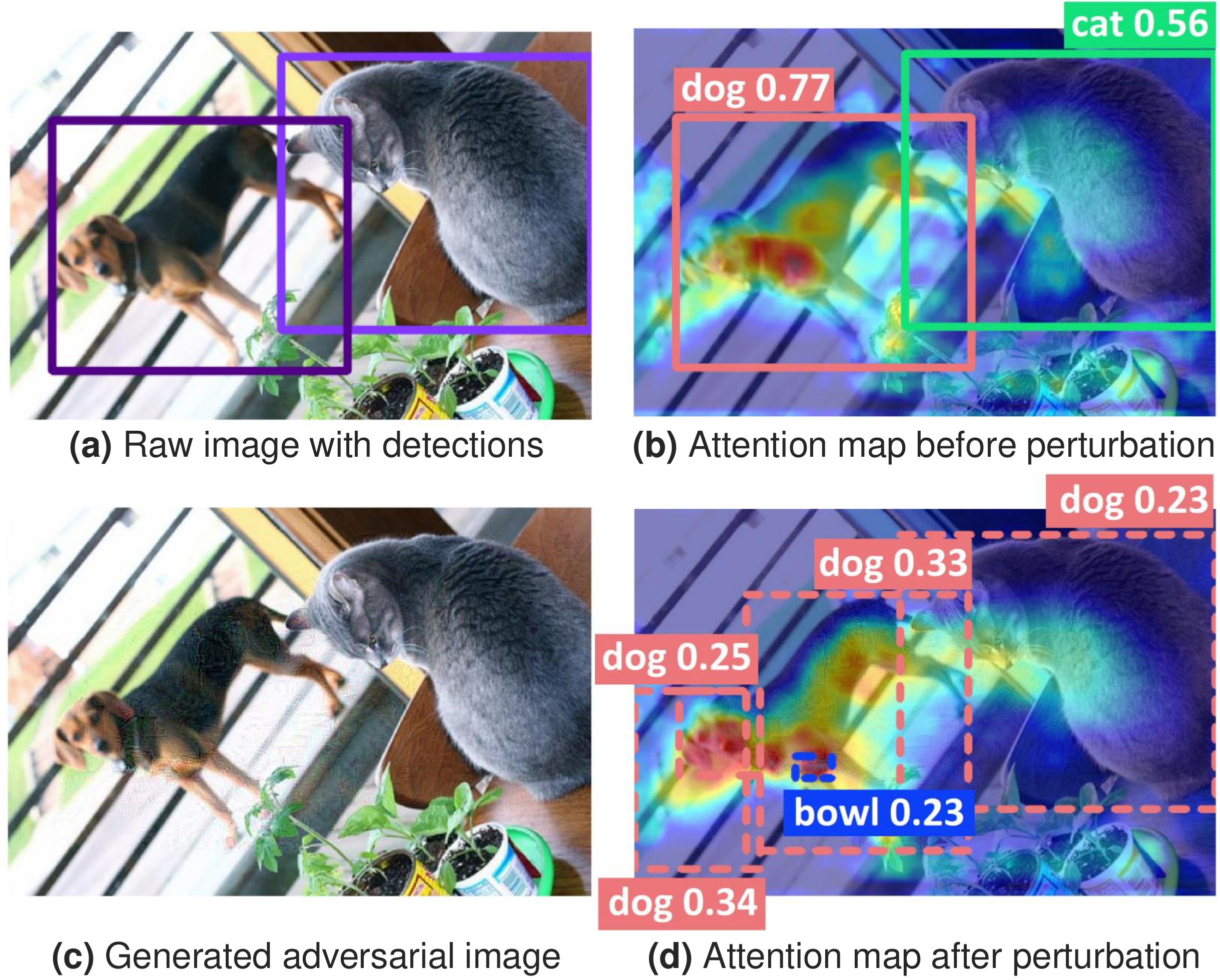}
    \caption{Visualization on attention maps before and after perturbation. The confidence scores of the detections on the attention regions are reduced after being attacked.}
    \vspace{7pt}
    \label{fig:stack_det_grad}
\end{figure}

\subsection{Success Rate of Adversarial Images to Models with Different Confidence Thresholds}
\label{sec:analyses_success}
Since different detectors are often set with different confidence thresholds, we analyze how many iterations our method takes to obtain the success attack rate of $100\%$, where all objects presented in the image are misdetected. Fig. \ref{fig:nonoverlap_overlap_attacks} (\textit{left}) shows the increasing effectiveness of noises added to the raw image over 90 iterations. In fact, with a confidence threshold of 0.50, the detector is unable to detect objects in the image; meanwhile, the detector with a confidence threshold of 0.25 can still detect objects, but the detections become inaccurate. starting from the 75$^{th}$ iteration. However, this particular image only illustrates results that the bounding boxes are not overlapped with each other. 

Fig. \ref{fig:nonoverlap_overlap_attacks} (\textit{right}) also shows the results where objects are overlapped with each other: the orange's bounding box is in the person's bounding box. Our method also obtains the success attack rate of $100\%$ to the model with the confidence threshold of 0.50. Nevertheless, this process takes about 580 iterations to completely fool the detector. 

\begin{table*}[t]
    \small
    \centering
    \resizebox{15.8cm}{!}{
    \begin{tabular}{c|cccccccc}
        \hline
        \textbf{\begin{tabular}[c]{@{}c@{}}Added Perturbation \end{tabular}} &
        \textbf{YOLOv8n} & \textbf{YOLOv8s} & \textbf{YOLOv8m} & \textbf{YOLOv8l} & \textbf{YOLOv8x} & \textbf{Faster R-CNN} & \textbf{RetinaNet} & \textbf{Swin-T}\\ \hline
        \textbf{None} (baseline) & 25.04  & 33.26 & 36.98  & 38.94  & 40.02 & 27.90 & 22.90 & 32.47 \\ \hline
        \textbf{YOLOv8n} & 0.06 & 18.12  & 25.19  & 28.25 & 29.52 & 13.57 & 10.69 & 17.22 \\ \hline
        \textbf{YOLOv8s} & 3.32  & 0.03 & 16.71   & 20.68 & 22.45 & 9.68&7.31&13.66\\ \hline
        \textbf{YOLOv8m} & 2.21  & 4.35 & 0.02 & 13.12 & 15.32 & 7.03 & 5.01 & 10.69\\ \hline
        \textbf{YOLOv8l} & 1.69 & 3.52 & 6.90  & 0.02 & 11.37 & 6.36 & 4.35 & 10.18 \\ \hline
        \rowcolor{Gray} \textbf{YOLOv8x} & 1.42 & 2.93 & 5.47  & 6.50 & 0.05 & 5.32 & 3.58 & 8.57\\ \hline
        \textbf{Faster R-CNN} & 3.86 & 6.96 & 10.51 & 13.09 & 13.96 & 0.10 & 0.60 & 12.70\\ \hline
        \textbf{RetinaNet} & 6.01 & 9.99 & 14.22 & 17.01 & 18.01 & 2.10 & 0.30 & 16.00\\ \hline
        \textbf{Swin-T} & 2.98 & 5.83 & 9.49 & 12.42 & 14.50 & 11.30 & 8.70 & 0.10 \\ \hline
    \end{tabular}}
    \caption{Cross-model transferability among commonly used detection models (in mAP) of various-sized YOLO's, Faster R-CNN, RetinaNet, and Swin Transformer, at confidence thresholds of $0.50$. Each model is evaluated on the MS COCO 2017 validation set as a baseline. Meanwhile, Alg. \ref{alg:adversarial_generation} best performs attacks on other models when generating adversarial perturbation against YOLOv8x.}
    \label{tab:transferability_COCO}
\end{table*}

\begin{table*}[hbt!]
    \small
    \centering
    \resizebox{15.8cm}{!}{
    \begin{tabular}{c|cccccccc}
    \hline
        \textbf{\begin{tabular}[c]{@{}c@{}}Added Perturbation \end{tabular}} &
        \textbf{YOLOv8n} & \textbf{YOLOv8s} & \textbf{YOLOv8m} & \textbf{YOLOv8l} & \textbf{YOLOv8x} & \textbf{Faster R-CNN} & \textbf{RetinaNet} & \textbf{Swin-T}\\ \hline
        \textbf{None} (baseline) & 45.15  & 54.45 & 60.80 & 63.47 & 64.00 & 46.13 & 49.54 & 53.35 \\ \hline
        \textbf{YOLOv8n} & 0.34 & 0.64 & 0.92 & 1.23 & 1.25 & 0.65 & 0.89 & 1.03 \\ \hline
        \textbf{YOLOv8s} & 0.36  & 0.39 & 0.80 & 1.07 & 1.08 & 0.60 & 0.86 & 0.95 \\ \hline
        \textbf{YOLOv8m} & 0.34  & 0.43 & 0.52 & 0.90 & 1.00 & 0.58 & 0.78 & 0.87 \\ \hline
        \textbf{YOLOv8l} & 0.35  & 0.48 & 0.65 & 0.70 & 0.88 & 0.49 & 0.62 & 0.75 \\ \hline
        \rowcolor{Gray} \textbf{YOLOv8x} & 0.31 & 0.45 & 0.61 & 0.66 & 0.72 & 0.41 & 0.58 & 0.67 \\ \hline
        \textbf{Faster R-CNN} & 5.13 & 9.04 & 16.02 & 18.51 & 19.75 & 0.09 & 1.42 & 17.23 \\ \hline
        \textbf{RetinaNet} & 8.84 & 13.94 & 21.47 & 23.89 & 25.57 & 1.97 & 0.12 & 21.97 \\ \hline
        \textbf{Swin-T} & 2.99 & 6.06 & 12.39 & 15.30 & 18.18 & 12.18 & 17.38 & 0.18 \\ \hline
    \end{tabular}}
    \caption{Cross-model transferability among commonly used detection models (in mAP) of various-sized YOLO's, Faster R-CNN, RetinaNet, and Swin Transformer, with confidence thresholds set to $0.50$. Each model is evaluated on the PASCAL VOC 2012 validation set as a baseline. Again, Alg. \ref{alg:adversarial_generation} best performs attacks on other models when generating adversarial perturbation against YOLOv8x.}
    \vspace{-11pt}
    \label{tab:transferability_PASCAL}
\end{table*}

\subsection{Attention of Detection Models}
\label{sec:anayses_attention}
To further explain our method, we analyze how the model's attention altered using Grad-CAM \cite{selvaraju2017grad}, as illustrated in Fig. \ref{fig:stack_det_grad}. Before being attacked (Fig. \ref{fig:stack_det_grad}a), the model can detect objects with high confidence scores, and its attention map (Fig. \ref{fig:stack_det_grad}b) accurately focuses on the areas presumed to contain objects. However, while performing Grad-CAM on perturbed images, the model fails to detect objects surpassing the confidence threshold (Fig. \ref{fig:stack_det_grad}d). Moreover, the model identifies the segmented regions, as visualized on attention maps, belonging to different classes. Also, as mentioned in Sec. \ref{sec:methodology_iterative_images}, our method strives to decrease the confidence scores of objects in each bounding box by determining the optimal noises, resulting in changes in the model's attention and, thereafter, its detection. Indeed, the attention map focuses on the same bounding boxes, and their intensities change since the confidence scores are reduced significantly, leading to misdetection. 


\textbf{Analysis Conclusions}: Our analyses allude that larger models might easily overcome adversarial attacks; however, this raises the concern of computing power while training these large-sized models with adversarial examples and deploying them for real-world applications. 

\vspace{-15pt}
\section{Experiments}
\label{sec:experiments}

We evaluate our method on MS COCO 2017 \cite{lin2014microsoft} and PASCAL VOC 2012 \cite{everingham2015pascal} datasets with other detection algorithms of different backbones, including validating with cross-model and cross-domain datasets and verifying their transferability to different backbones and consistency with different detection algorithms. The experiments are conducted as follows: (1) generating adversarial images against one detector, then (2) perturbing other detectors using those images without prior knowledge about the models.

\subsection{Cross-Model Validation}
\label{sec:cross-model-validation}
We use pre-trained models (YOLOv8, Faster-RCNN, RetinaNet, Swin Transformer) trained on MS COCO 2017 and generate adversarial examples for each model on the validation set of MS COCO 2017. The adversarial examples generated by one model are evaluated by others, including itself. Tab. \ref{tab:transferability_COCO} shows that models are fooled by adversarial images generated by themselves, in which these images include knowledge of that model: the most optimal (best) perturbation to make that specific model misdetect. 

The results also show that the larger-sized models generate adversarial examples that are more effective against smaller ones. Notably, also from Tab. \ref{tab:transferability_COCO}, our method best performs when testing its adversarial examples (against YOLOv8x) on other models since it produces more generalized noises affecting other models. Excluding attacking itself, these adversarial images best attack YOLOv8s and worst attack Swin-T with 91.19\% (dropping the model's mAP from 33.26 to 2.93) and 73.61\% (from 32.47 down to 8.57) success attack rates, respectively. 

\subsection{Cross-Domain Datasets Validation}
\label{sec:croos-domain-validation}
To verify the generality of our attacking method, we also conduct experiments in which models are trained on one dataset and evaluated on another dataset. As presented in Sec. \ref{sec:cross-model-validation}, models are trained on MS COCO 2017, and adversarial examples are also generated from MS COCO 2017. Tab. \ref{tab:transferability_PASCAL} shows that transferability is robust on another dataset, where pre-trained models on MS COCO 2017 are tested with adversarial examples generated from the validation set of PASCAL VOC 2012.

Similar to Tab. \ref{tab:transferability_COCO}, our method again shows its best performance when testing its adversarial examples (against YOLOv8x) on other models, where these adversarial examples best attack YOLOv8n and worst attack Swin-T with 99.31\% (from 45.15 down to 0.31) and 73.61\% (from 53.35 down to 0.67) success attack rates, respectively. Moreover, generated adversarial examples against YOLOv8x on the PASCAL VOC 2012 validation set even outperform those generated on the MS COCO 2017 validation set; indeed, they achieve the average success attack rates of 99\% compared to 86.6\% of average success attack rate.

\begin{figure*}[t]
    \centering
    \includegraphics[width=1.00\linewidth]{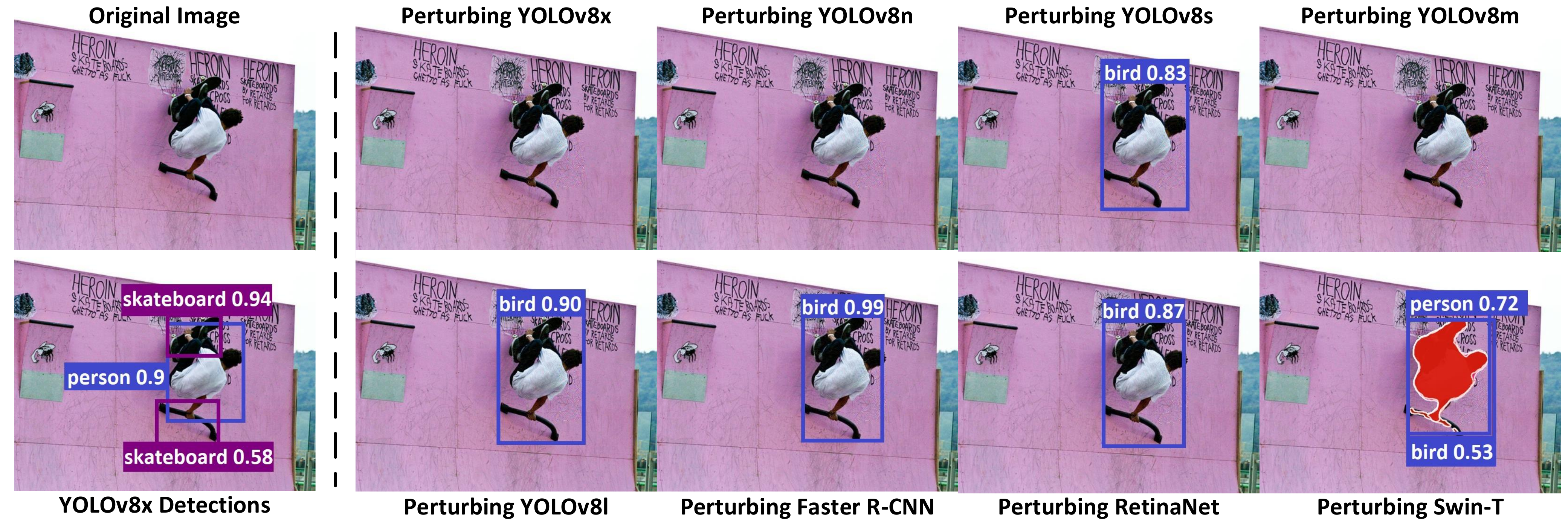}
    \caption{Qualitative results of adversarial images against YOLOv8x that perturbs other detection models, including YOLO's versions, Faster R-CNN, RetinaNet, and Swin Transformer, at confidence thresholds of $0.50$.} 
    \vspace{-10pt}
    \label{fig:qualitative_results}
\end{figure*}

\subsection{Attack Transferability}
\label{sec:experiments_transferability}
Furthermore, we compare our methods with DAG \cite{xie2017adversarial} regarding the transferability to other backbones: adversarial images generated against a different backbone are used to attack detectors with ResNet-50 as backbones. We used the images (from the PASCAL VOC dataset) generated against YOLOv8x to perturb Faster R-CNN, RetinaNet, and Swin Transformer. As shown in Tab. \ref{tab:transferability_backbones}, we still achieve a success attack rate of 80.44\%, 84.37\%, and 73.61\%, respectively; meanwhile, DAG only achieved 16.23\% while performing the same task. 

\begin{table}[t]
    \small
    \centering
    \resizebox{7.6cm}{!}{
    \begin{tabular}{c| M{1.1cm} M{1.1cm} M{1.1cm} | M{1.1cm}}
        \hline
        {} & \multicolumn{4}{c}{\textbf{ResNet-50 Backbone}} \\
        \hline
        {} & Faster R-CNN & RetinaNet & Swin-T & R-FCN \par -RN50 \\
        \hline
        Baseline & 27.20 & 22.90 & 32.47 & 76.40 \\
        \hline
        DAG & - & - & - & 63.93 \\
        \hline 
        \rowcolor{Gray} Ours & 5.32 & 3.58 & 8.57 & - \\
        \hline
        Succ. Rate & 80.44\% & 84.37\% & 73.61\% & 16.32\%\\
        \hline 
    \end{tabular}}
    \caption{Comparisons of success attack rates between DAG \cite{xie2017adversarial} and our proposed method on detection models with ResNet-50 backbone.}
    \vspace{6pt}
    \label{tab:transferability_backbones}
\end{table}

\subsection{Consistency with Detection Algorithms}
\label{sec:experiments_consistency}
Also, to see how consistent Alg. \ref{alg:adversarial_generation} performs with different detection algorithms, we experiment it on both one-stage and two-stage detection algorithms and compare our results with DAG \cite{xie2017adversarial} and UEA \cite{DBLP:conf/ijcai/WeiLCC19}, as depicted in Tab. \ref{tab:consistency_algorithms}. All three methods provide high results (above 90\%) on one-stage detection methods; however, the performances of DAG and UEA drop when performing adversarial attacks on two-stage detection methods, while our proposed technique can still maintain a consistent success attack rate of 92.47\% compared to 93.25\% from one-stage methods. 

\begin{table}[t]
    \small
    \centering
    \resizebox{7.6cm}{!}{
    \begin{tabular}{c| M{0.67cm} M{0.67cm} M{0.81cm} | M{0.55cm} M{0.67cm} M{0.67cm} }
        \hline
        {} & \multicolumn{3}{c}{\textbf{One-Stage}} & \multicolumn{3}{c}{\textbf{Two-Stage}} \\
        \hline
        Baseline & 68.00 & 68.00 & 25.04 & 70.10 & 70.10 & 27.90 \\
        \hline
        DAG & 5.00 & - & - & 64.00 & - & - \\
        UEA & - & 5.00 & - & - & 20.00 & - \\
        \hline 
        \rowcolor{Gray} Ours & - & - & 1.69 & - & - & 2.10 \\
        \hline
        Succ. Rate & 92.65\% & 92.65\% & 93.25\% & 8.70\% & 71.47\% & 92.47\% \\
        \hline 
    \end{tabular}}
    \caption{Success attack rates between DAG \cite{xie2017adversarial}, UEA \cite{DBLP:conf/ijcai/WeiLCC19}, and our method on one-stage and two-stage detection algorithms.}
    \vspace{8pt}
    \label{tab:consistency_algorithms}
\end{table}

\subsection{Qualitative Results}
From Tab. \ref{tab:transferability_COCO} and Tab. \ref{tab:transferability_PASCAL}, we conclude that adversarial images generated against YOLOv8x maintain the best overall transferability and consistency of attacks to other models. As shown in Fig. \ref{fig:qualitative_results}, the qualitative results of a perturbed image against YOLOv8x can make other detection models misdetect. Fig. \ref{fig:qualitative_results} also shows that the perturbation amount is imperceptible, the stable transferability to other backbones, and the consistency with one-stage and two-stage methods, restating our key properties in Tab. \ref{tab:key_properties}.

\subsection{Discussions}
Our cross-model validation experiments demonstrate the strong transferability of adversarial examples across diverse detection architectures. Adversarial images crafted against YOLOv8x effectively misled other YOLOv8 variants, as well as models like Faster R-CNN, RetinaNet, and Swin Transformer, achieving high success rates. Notably, larger models, such as YOLOv8x, not only demonstrated greater robustness but also generated adversarial examples that generalized better to other models. This trend suggests that larger models’ architectural complexity enables them to produce perturbations that impact shared features across different backbones.

Cross-domain validations further support the generalizability of our method. Adversarial examples generated on the MS COCO 2017 dataset remained effective when tested on PASCAL VOC 2012, achieving success rates comparable to in-domain experiments. These results underline the robustness of our perturbation approach, which leverages model-agnostic loss gradients to craft transferable adversarial examples. This ability to maintain high efficacy across datasets enhances the practicality of our method for black-box attack scenarios, where access to target model specifics is limited.

The transferability of adversarial examples to different backbones also highlights the adaptability of our approach. Using adversarial examples generated against YOLOv8x, we observed consistent attack success rates on models with ResNet-50 backbones, such as Faster R-CNN and RetinaNet, and even on transformer-based models like Swin Transformer. These findings indicate that our method effectively exploits fundamental vulnerabilities in object detection pipelines, regardless of the underlying network architecture.

Our experiments also confirm the consistency of our method across one-stage and two-stage detection algorithms. While prior methods like DAG and UEA showed a drop in performance on two-stage detectors, our technique maintained high success rates across both categories. This consistency is attributed to the iterative perturbation approach, which targets bounding box regions while controlling distortion, ensuring applicability across different detection paradigms.

Qualitative results and visual analyses provide further evidence of our method’s efficacy. Grad-CAM visualizations reveal how adversarial perturbations alter model attention, reducing confidence scores for objects in bounding boxes and eventually leading to misdetections. Additionally, the perturbations remain imperceptible to human observers, striking an effective balance between visual fidelity and attack performance. These properties make our approach suitable for real-world applications where stealth is essential.

Despite these strengths, our method encounters challenges in scenarios with overlapping bounding boxes, requiring more iterations and distortion to achieve similar success rates. Addressing these limitations through advanced perturbation strategies or adaptive adversarial training could enhance the robustness of future detection systems. Furthermore, exploring domain adaptation techniques may improve cross-domain transferability even further.

\vspace{-10pt}
\section{Conclusions}
This paper presents a distortion-aware adversarial attack technique on bounding boxes of state-of-the-art object detectors by leveraging target-attacked pixel gradient ascents. By knowing the gradient ascents of those pixels, we iteratively add the perturbation amount to the original image's masked regions until the success attack rate or distortion threshold is obtained or until the detector no longer recognizes the presented objects. To verify the effectiveness of the proposed method, we evaluate our approach on MS COCO 2017 and PASCAL VOC 2012 datasets and achieve success attack rates of up to $100$\% and $98$\%, respectively. Also, through validating cross-model transferability, we prove that our method can perform \textit{black-box attacks} when generating primary adversarial images on YOLOv8x. As the original motivation of our work, we propose this method to expose the vulnerabilities in neural networks and facilitate building more reliable detection models under adversaries. We reserve the task of improving the model's robustness for future works. Upon social goods, we make our source code available to encourage others to build defense methods for this attack mechanism.

\vspace{-5pt}
\bibliographystyle{apalike}
{\small \bibliography{08_references}}

\begin{thebibliography}{}

\bibitem[Alaifari et~al., 2018]{alaifariadef}
Alaifari, R., Alberti, G.~S., and Gauksson, T. (2018).
\newblock Adef: an iterative algorithm to construct adversarial deformations.
\newblock In {\em International Conference on Learning Representations}.

\bibitem[Bochkovskiy et~al., 2020]{bochkovskiyyolov4}
Bochkovskiy, A., Wang, C.-Y., and Liao, H.-Y.~M. (2020).
\newblock Yolov4: Optimal speed and accuracy of object detection.

\bibitem[Carlini and Wagner, 2017]{carlini2017towards}
Carlini, N. and Wagner, D. (2017).
\newblock Towards evaluating the robustness of neural networks.
\newblock In {\em 2017 ieee symposium on security and privacy (sp)}, pages 39--57. Ieee.

\bibitem[Chen et~al., 2018]{chen2018ead}
Chen, P.-Y., Sharma, Y., Zhang, H., Yi, J., and Hsieh, C.-J. (2018).
\newblock Ead: elastic-net attacks to deep neural networks via adversarial examples.
\newblock In {\em Proceedings of the AAAI conference on artificial intelligence}, volume~32.

\bibitem[Dang et~al., 2023]{dang2023multiplanar}
Dang, T., Nguyen, K., and Huber, M. (2023).
\newblock Multiplanar self-calibration for mobile cobot 3d object manipulation using 2d detectors and depth estimation.
\newblock In {\em 2023 IEEE/RSJ International Conference on Intelligent Robots and Systems (IROS)}, pages 1782--1788. IEEE.

\bibitem[Dang et~al., 2024]{dang2024v3d}
Dang, T., Nguyen, K., and Huber, M. (2024).
\newblock V3d-slam: Robust rgb-d slam in dynamic environments with 3d semantic geometry voting.
\newblock {\em arXiv preprint arXiv:2410.12068}.

\bibitem[Du et~al., 2022]{du_physical_2022}
Du, A., Chen, B., Chin, T.-J., Law, Y.~W., Sasdelli, M., Rajasegaran, R., and Campbell, D. (2022).
\newblock Physical {Adversarial} {Attacks} on an {Aerial} {Imagery} {Object} {Detector}.
\newblock pages 1796--1806.

\bibitem[Everingham et~al., 2015]{everingham2015pascal}
Everingham, M., Eslami, S.~A., Van~Gool, L., Williams, C.~K., Winn, J., and Zisserman, A. (2015).
\newblock The pascal visual object classes challenge: A retrospective.
\newblock {\em International journal of computer vision}, 111:98--136.

\bibitem[Goodfellow et~al., 2015]{DBLP:journals/corr/GoodfellowSS14}
Goodfellow, I.~J., Shlens, J., and Szegedy, C. (2015).
\newblock Explaining and harnessing adversarial examples.
\newblock In Bengio, Y. and LeCun, Y., editors, {\em 3rd International Conference on Learning Representations, {ICLR} 2015, San Diego, CA, USA, May 7-9, 2015, Conference Track Proceedings}.

\bibitem[Im~Choi and Tian, 2022]{im2022adversarial}
Im~Choi, J. and Tian, Q. (2022).
\newblock Adversarial attack and defense of yolo detectors in autonomous driving scenarios.
\newblock In {\em 2022 IEEE Intelligent Vehicles Symposium (IV)}, pages 1011--1017. IEEE.

\bibitem[Jocher et~al., 2023]{Jocher_YOLO_by_Ultralytics_2023}
Jocher, G., Chaurasia, A., and Qiu, J. (2023).
\newblock {YOLO by Ultralytics}.

\bibitem[Kurakin et~al., 2018]{kurakin2018adversarial}
Kurakin, A., Goodfellow, I.~J., and Bengio, S. (2018).
\newblock Adversarial examples in the physical world.
\newblock In {\em Artificial intelligence safety and security}, pages 99--112. Chapman and Hall/CRC.

\bibitem[Lin et~al., 2017]{lin2017focal}
Lin, T.-Y., Goyal, P., Girshick, R., He, K., and Doll{\'a}r, P. (2017).
\newblock Focal loss for dense object detection.
\newblock In {\em Proceedings of the IEEE international conference on computer vision}, pages 2980--2988.

\bibitem[Lin et~al., 2014]{lin2014microsoft}
Lin, T.-Y., Maire, M., Belongie, S., Hays, J., Perona, P., Ramanan, D., Doll{\'a}r, P., and Zitnick, C.~L. (2014).
\newblock Microsoft coco: Common objects in context.
\newblock In {\em Computer Vision--ECCV 2014: 13th European Conference, Zurich, Switzerland, September 6-12, 2014, Proceedings, Part V 13}, pages 740--755. Springer.

\bibitem[Lindeberg, 2012]{lindeberg2012scale}
Lindeberg, T. (2012).
\newblock Scale invariant feature transform.

\bibitem[Liu et~al., 2019]{DBLP:conf/aaai/LiuYLSCL19}
Liu, X., Yang, H., Liu, Z., Song, L., Chen, Y., and Li, H. (2019).
\newblock {DPATCH:} an adversarial patch attack on object detectors.
\newblock In Espinoza, H., h{\'{E}}igeartaigh, S.~{\'{O}}., Huang, X., Hern{\'{a}}ndez{-}Orallo, J., and Castillo{-}Effen, M., editors, {\em Workshop on Artificial Intelligence Safety 2019 co-located with the Thirty-Third {AAAI} Conference on Artificial Intelligence 2019 (AAAI-19), Honolulu, Hawaii, January 27, 2019}, volume 2301 of {\em {CEUR} Workshop Proceedings}. CEUR-WS.org.

\bibitem[Liu et~al., 2021]{liu2021swin}
Liu, Z., Lin, Y., Cao, Y., Hu, H., Wei, Y., Zhang, Z., Lin, S., and Guo, B. (2021).
\newblock Swin transformer: Hierarchical vision transformer using shifted windows.
\newblock In {\em Proceedings of the IEEE/CVF international conference on computer vision}, pages 10012--10022.

\bibitem[Lu et~al., 2017]{lu_adversarial_2017}
Lu, J., Sibai, H., and Fabry, E. (2017).
\newblock Adversarial {Examples} that {Fool} {Detectors}.
\newblock arXiv:1712.02494 [cs].

\bibitem[Lu, 2019]{lu_level_2019}
Lu, Y. (2019).
\newblock The {Level} {Weighted} {Structural} {Similarity} {Loss}: {A} {Step} {Away} from {MSE}.
\newblock {\em Proceedings of the AAAI Conference on Artificial Intelligence}, 33(01):9989--9990.
\newblock Number: 01.

\bibitem[Madry et~al., 2018]{DBLP:conf/iclr/MadryMSTV18}
Madry, A., Makelov, A., Schmidt, L., Tsipras, D., and Vladu, A. (2018).
\newblock Towards deep learning models resistant to adversarial attacks.
\newblock In {\em 6th International Conference on Learning Representations, {ICLR} 2018, Vancouver, BC, Canada, April 30 - May 3, 2018, Conference Track Proceedings}. OpenReview.net.

\bibitem[Moosavi-Dezfooli et~al., 2016]{moosavi-dezfooli_deepfool_2016}
Moosavi-Dezfooli, S.-M., Fawzi, A., and Frossard, P. (2016).
\newblock {DeepFool}: {A} {Simple} and {Accurate} {Method} to {Fool} {Deep} {Neural} {Networks}.
\newblock pages 2574--2582.

\bibitem[Nguyen et~al., 2024a]{nguyen2024real}
Nguyen, K., Dang, T., and Huber, M. (2024a).
\newblock Real-time 3d semantic scene perception for egocentric robots with binocular vision.
\newblock {\em arXiv preprint arXiv:2402.11872}.

\bibitem[Nguyen et~al., 2024b]{nguyen2024volumetric}
Nguyen, K., Dang, T., and Huber, M. (2024b).
\newblock Volumetric mapping with panoptic refinement via kernel density estimation for mobile robots.
\newblock {\em arXiv preprint arXiv:2412.11241}.

\bibitem[Puccetti et~al., 2023]{puccetti2023efficacy}
Puccetti, T., Zoppi, T., and Ceccarelli, A. (2023).
\newblock On the efficacy of metrics to describe adversarial attacks.
\newblock {\em arXiv preprint arXiv:2301.13028}.

\bibitem[Ren et~al., 2015]{ren2015faster}
Ren, S., He, K., Girshick, R., and Sun, J. (2015).
\newblock Faster r-cnn: Towards real-time object detection with region proposal networks.
\newblock {\em Advances in neural information processing systems}, 28.

\bibitem[Rublee et~al., 2011]{rublee2011orb}
Rublee, E., Rabaud, V., Konolige, K., and Bradski, G. (2011).
\newblock Orb: An efficient alternative to sift or surf.
\newblock In {\em 2011 International conference on computer vision}, pages 2564--2571. Ieee.

\bibitem[Selvaraju et~al., 2017]{selvaraju2017grad}
Selvaraju, R.~R., Cogswell, M., Das, A., Vedantam, R., Parikh, D., and Batra, D. (2017).
\newblock Grad-cam: Visual explanations from deep networks via gradient-based localization.
\newblock In {\em Proceedings of the IEEE international conference on computer vision}, pages 618--626.

\bibitem[Song et~al., 2018]{song2018physical}
Song, D., Eykholt, K., Evtimov, I., Fernandes, E., Li, B., Rahmati, A., Tramer, F., Prakash, A., and Kohno, T. (2018).
\newblock Physical adversarial examples for object detectors.
\newblock In {\em 12th USENIX workshop on offensive technologies (WOOT 18)}.

\bibitem[Wang et~al., 2004]{wang2004image}
Wang, Z., Bovik, A.~C., Sheikh, H.~R., and Simoncelli, E.~P. (2004).
\newblock Image quality assessment: from error visibility to structural similarity.
\newblock {\em IEEE transactions on image processing}, 13(4):600--612.

\bibitem[Wei et~al., 2019]{DBLP:conf/ijcai/WeiLCC19}
Wei, X., Liang, S., Chen, N., and Cao, X. (2019).
\newblock Transferable adversarial attacks for image and video object detection.
\newblock In Kraus, S., editor, {\em Proceedings of the Twenty-Eighth International Joint Conference on Artificial Intelligence, {IJCAI} 2019, Macao, China, August 10-16, 2019}, pages 954--960. ijcai.org.

\bibitem[Xie et~al., 2017]{xie2017adversarial}
Xie, C., Wang, J., Zhang, Z., Zhou, Y., Xie, L., and Yuille, A. (2017).
\newblock Adversarial examples for semantic segmentation and object detection.
\newblock In {\em Proceedings of the IEEE international conference on computer vision}, pages 1369--1378.

\end{thebibliography}

\end{document}